\documentclass[10pt,twocolumn,letterpaper]{article}

\usepackage[pagenumbers]{cvpr} 

\definecolor{cvprblue}{rgb}{0.21,0.49,0.74}
\usepackage[pagebackref,breaklinks,colorlinks,allcolors=cvprblue]{hyperref}
\usepackage{algorithm}
\usepackage{algorithmic}

\def\paperID{1693} 
\def\confName{CVPR}
\def\confYear{2026}

\title{In-Context Audio Control of Video Diffusion Transformers}

\author{
Wenze Liu$^{1*}$ \quad Weicai Ye$^{2\dag}$ \quad Minghong Cai$^{1}$ \quad Quande Liu$^{2}$ \quad Xintao Wang$^{2}$ \quad Xiangyu Yue$^{1\dag}$ \\
$^{1}$MMLab, The Chinese University of Hong Kong \quad $^{2}$Kling Team, Kuaishou Technology
}

\begin{document}
\maketitle
\def\thefootnote{*}\footnotetext{Work done at Kuaishou Technology.}
\def\thefootnote{\dag}\footnotetext{Corresponding authors.}
\def\thefootnote{\arabic{footnote}}
\begin{abstract}
Recent advancements in video generation have seen a shift towards unified, transformer-based foundation models that can handle multiple conditional inputs in-context. However, these models have primarily focused on modalities like text, images, and depth maps, while strictly time-synchronous signals like audio have been underexplored. This paper introduces In-Context Audio Control of video diffusion transformers (ICAC), a framework that investigates the integration of audio signals for speech-driven video generation within a unified full-attention architecture, akin to FullDiT. We systematically explore three distinct mechanisms for injecting audio conditions: standard cross-attention, 2D self-attention, and unified 3D self-attention. Our findings reveal that while 3D attention offers the highest potential for capturing spatio-temporal audio-visual correlations, it presents significant training challenges. To overcome this, we propose a Masked 3D Attention mechanism that constrains the attention pattern to enforce temporal alignment, enabling stable training and superior performance. Our experiments demonstrate that this approach achieves strong lip synchronization and video quality, conditioned on an audio stream and reference images.
\end{abstract}    
\section{Introduction}
\label{sec:intro}

The landscape of generative artificial intelligence has been reshaped by the remarkable strides in video synthesis, a domain profoundly influenced by the advent of diffusion models \cite{sohl2015deep,ho2020denoising,song2019generative}. In recent years, a significant architectural evolution has been underway, marking a paradigm shift from the long-dominant U-Net architectures \cite{ho2022imagen,blattmann2023stable,chen2023videocrafter1,jiang2025fairgen} to the more scalable and powerful large-scale diffusion transformers (DiTs) \cite{hong2022cogvideo,yang2024cogvideox,videoworldsimulators2024,ma2024latte,kong2024hunyuanvideo,wan2025wan,ju2025fulldit}. A particularly exciting frontier within this new paradigm is the development of unified foundation models, designed to process a diverse array of conditional inputs within a single, coherent framework. Capitalizing on the exceptional sequence modeling capabilities of transformers, models like FullDiT \cite{ju2025fulldit} have showcased impressive success in generating video from a heterogeneous mix of inputs including text, reference images, camera paths, and depth maps by tokenizing all modalities into a unified sequence and processing them with a full self-attention mechanism. This `in-context' generation approach promises unprecedented flexibility and scalability.

However, despite this progress, a critical gap persists in the ability of these unified models to handle conditioning signals that demand strict, moment-to-moment temporal synchronization with the generated video. The most prominent example of such a signal is speech audio. Audio-driven video generation, particularly the synthesis of talking heads, is a task of immense practical importance, with applications spanning from creating digital avatars and virtual assistants to enhancing film production and accessibility tools. While previous works \cite{wei2025mocha,cui2025hallo3,ji2025sonic,jiang2024loopy} have produced impressive and high-fidelity results, they typically rely on specialized architectures. These models often employ separate processing branches, dedicated audio encoders, or custom-designed modules to inject the audio information into a video synthesis backbone. Such specialized designs, while effective for their specific task, are fundamentally at odds with the modality-agnostic philosophy of a unified, full-attention foundation model, hindering their extensibility and scalability.

This paper directly confronts the challenge of natively integrating a synchronous, dynamic signal like audio into an in-context video generation framework. We introduce ICAC, a model and a systematic exploration of audio injection architectures within a FullDiT-like transformer. Our investigation compares three distinct strategies for conditioning the video generation process on an audio signal: a baseline cross-attention mechanism, a more integrated 2D self-attention approach (with or without updating audio features), and a fully unified spatio-temporal 3D self-attention architecture. Through empirical analysis, we uncover a crucial trade-off: as the level of integration deepens, the training process becomes progressively more challenging, with the fully unified 3D model failing to converge altogether. We diagnose this instability as a consequence of the vast feature disparity and the strict synchronization requirements between the audio and video modalities; the powerful, static features from the identity reference image tend to dominate the learning process, preventing the model from effectively learning the subtle, dynamic cues from the audio signal.

To resolve this critical issue, we propose and develop a Masked 3D Attention mechanism. This approach constrains the powerful self-attention mechanism by encouraging video and audio feature tokens to primarily attend to each other within the same temporal frame. This enforces a strong inductive bias for learning instantaneous audio-visual correlations—essential for accurate lip-sync, while preserving the model's ability to capture broader contextual relationships. For computational feasibility in a large-scale transformer, we also develop an efficient implementation leveraging Flash Attention. Our masked attention strategy successfully resolves the convergence issue, and with it enabled, we observe a clear performance hierarchy: the deeper the integration of the audio signal, the better the final result. This confirms that our Masked 3D Attention model is the most effective architecture, successfully unlocking the full potential of a unified approach. Building on this powerful foundation, our final ICAC framework can generate high-quality, audio-driven videos conditioned on diverse reference images.

In summary, our primary contributions are threefold:
\begin{itemize}
\item We provide a systematic comparison and analysis of three distinct strategies for injecting audio into a unified video transformer, highlighting the inherent stability challenges of deep integration.
\item We devise a Masked 3D Attention mechanism that resolves the training instability of a fully spatio-temporal approach, enabling it to achieve superior performance by effectively balancing global context with local synchronization.
\item We demonstrate that our resulting framework, ICAC, is capable of generating controllable and interactive talking character videos from audio and image references, advancing the state of the art in unified, multi-modal video synthesis.
\end{itemize}
\section{Related Work}
\label{sec:related}

\subsection{Audio-Driven Video Generation}

The synthesis of photorealistic talking characters from an audio signal has been a long-standing and pivotal goal in computer vision and graphics. Early approaches often adopted a structured, multi-stage pipeline. Models like SadTalker \cite{zhang2023sadtalker} and AniPortrait \cite{wei2024aniportrait} first predict intermediate representations, such as 3D facial meshes or 2D landmarks, from the audio. A separate synthesis network then renders the video conditioned on these explicit movements. While this provides control, the ultimate realism can be limited by the expressiveness of the intermediate geometry, which may not capture the full range of subtle facial dynamics and natural head motions. The rise of diffusion models has spurred a shift towards end-to-end methods that generate video directly from audio, resulting in more natural motion and higher visual fidelity. State-of-the-art U-Net-based models like EMO \cite{tian2024emo} and Hallo3 \cite{cui2025hallo3} produce highly expressive results but often rely on specialized modules or auxiliary conditions to maintain identity and temporal coherence. More recently, the field has explored Diffusion Transformers (DiTs) for this task. MoCha \cite{wei2025mocha} marked a significant advance by training a DiT with a novel speech-video window attention mechanism to enforce local temporal alignment. Similarly, Loopy \cite{jiang2024loopy} proposed specialized temporal modules to capture long-range motion dependencies, and EchoMimicV2 \cite{meng2025echomimicv2} extended control to include semi-body animations. While these methods are undeniably powerful, their architectural designs are highly specialized and fine-tuned for the talking-head generation task. This specialization makes them difficult to integrate into or adapt for a general-purpose, multi-modal, in-context generation framework, which is the central focus of our work.

\subsection{Unified Multi-Task Video Generation}

In parallel to advancements in specific generation tasks, the concept of building a single, versatile foundation model capable of handling a multitude of generation tasks has gained significant traction. In the image domain, an early and influential approach has been the use of adapter-based modules. Frameworks like ControlNet \cite{zhang2023adding} and T2I-Adapter \cite{mou2024t2i} introduce lightweight, trainable modules that inject new forms of spatial control (\eg, edge maps, human pose, depth) into a large, pre-trained, and frozen text-to-image diffusion model. While this approach is parameter-efficient and effective for adding a single new condition, it can suffer from branch conflicts and parameter redundancy when multiple adapters are naively combined. A more recent and arguably more elegant paradigm seeks to unify all conditional inputs into a single, shared representation space. OminiControl \cite{tan2024ominicontrol} provides a compelling alternative by tokenizing all conditional inputs into a single sequence. This sequence is then processed by a FLUX-style \cite{flux2024} transformer architecture equipped with multi-modal attention mechanisms \cite{pan2020multi}, allowing the model to learn complex inter-modal relationships within a unified computational graph. FullDiT \cite{ju2025fulldit} successfully extends this in-context generation paradigm to the more complex domain of video generation. It demonstrates that diverse conditions such as a reference identity image, a camera motion trajectory, and a depth map sequence can all be tokenized and concatenated to form one contiguous sequence. This sequence is then processed by a shared full self-attention mechanism, which learns to interpret the context and generate a coherent video. This unified approach inherently eliminates the need for separate adapters and resolves potential condition conflicts. Our work is built directly upon this powerful and flexible philosophy. We aim to bridge a critical gap by exploring, for the first time, how a high-frequency, continuous signal that demands strict temporal synchronization like speech audio can be effectively and seamlessly integrated into such a unified full-attention framework.
\section{Preliminary}
\label{sec:preliminary}

\subsection{Latent Video Diffusion Model}

To handle the high dimensionality of video data, video generative models operate within a compressed latent space, a common and effective strategy for generative models. The first step involves a pre-trained spatio-temporal Variational Autoencoder (3D-VAE). Its encoder takes a raw video clip from the pixel space and maps it to a much lower-dimensional latent representation. Conversely, the decoder can transform a latent representation back into a high-fidelity video clip. This compression makes the subsequent generative modeling task computationally feasible.

The core of our generative process is a diffusion model, for which we use the flow matching formulation \cite{lipman2022flow,liu2022flow}, a simple and efficient variant. Given a clean data sample $\boldsymbol{x}_1$ (which is the latent representation of a real video clip from the 3D-VAE) and a random noise sample $\boldsymbol{x}_0 \sim \mathcal{N}(\boldsymbol{0},\boldsymbol{I})$ of the same dimensionality, flow matching defines a continuous path between them. An intermediate latent state $\boldsymbol{x}_t$ is constructed by linearly interpolating between the noise and the data for a continuous timestep $t \in [0,1]$:
\begin{equation}
\label{eq:flow_matching_path}
\boldsymbol{x}_t = (1-t)\boldsymbol{x}_0 + t\boldsymbol{x}_1
\end{equation}

The model is then trained to predict the velocity of this flow, which is the rate of change along the path. The target estimation is simply defined as the difference between the clean data and the noise $\boldsymbol{x}_1 - \boldsymbol{x}_0$. The training objective is to minimize the $\ell_2$ distance between the model's prediction and the target:
\begin{equation}
\label{eq:flow_matching_loss}
\mathcal{L}(\theta) = \mathbb{E}_{\boldsymbol{x}_0,\boldsymbol{x}_1,t} \|\boldsymbol{v}_\theta(\boldsymbol{x}_t, t, \boldsymbol{c}) - (\boldsymbol{x}_1 - \boldsymbol{x}_0)\|_2^2
\end{equation}
where $\boldsymbol{c}$ denotes any conditional inputs, such as text, images, and audio.

We use a diffusion transformer (DiT) as the core generative model. The input to DiT is first divided into a sequence of non-overlapping spatio-temporal patches or tokens. These tokens along with other conditional tokens are then processed by a series of transformer blocks, which use self-attention to model the complex relationships between all parts of the video and the conditioning signals. The final output is a sequence of tokens that represents the predicted velocity for each corresponding input patch.

\subsection{In-context Video Generation}

Our work builds upon the paradigm of \textit{in-context video generation}, as pioneered by models like FullDiT \cite{ju2025fulldit}. This approach aims to create a single, multi-task generative foundation model that processes various conditional inputs through a unified full-attention framework, rather than relying on separate, task-specific adapter modules.

The core architectural principle is to convert all inputs into a single, coherent sequential representation. The process begins by tokenizing the noisy video latent $\boldsymbol{x}_t$ and each condition such as camera parameters, identity reference images, and depth maps into its own sequence of tokens. These individual sequences are then simply concatenated into one long, unified sequence. This combined sequence is fed into a series of transformer blocks. Within these blocks, a unified full 3D self-attention mechanism is applied across the entire sequence, allowing for joint modeling and deep interaction between all conditions across both spatial and temporal dimensions. This elegant approach avoids the potential branch conflicts and parameter redundancy common in adapter-based methods and facilitates a more thorough and flexible fusion of information from different modalities. In the following sections, we extend this paradigm by exploring how to effectively inject a synchronized, continuous conditional signal like audio into the FullDiT framework.
\section{Method}
\label{sec:method}

\begin{figure}[t!]
\begin{center}
\includegraphics[width=\linewidth]{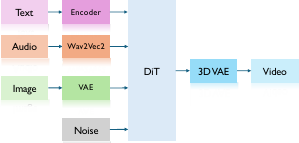}
\end{center}
\caption{The overall architecture of ICAC. It takes text, audio, image, and noise as inputs. The text is processed by a text encoder, the audio by Wav2Vec2, and the image by a VAE to be encoded, respectively. These encoded conditional inputs, along with the noise, are fed into a DiT model. Finally, the output from the DiT is decoded by a 3D VAE to generate the final Video.}
\label{fig:architecture}
\end{figure}

\begin{figure*}[t!]
\begin{center}
\includegraphics[width=\linewidth]{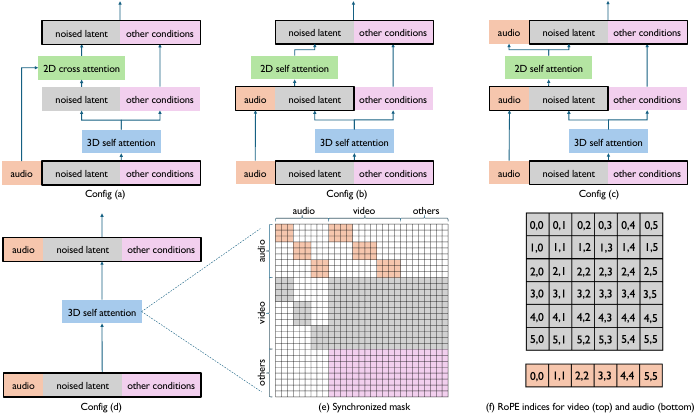}
\end{center}
\caption{Conceptual visualization of the compared attention configurations including (a) 2D cross-attention, (b) 2D self-attention without updating audio features, (c) 2D self-attention and (d) (masked) 3D attention. On the right of (d), we show the mask pattern of masked 3D attention (e), consisting of video-to-video attention (full), audio-to-audio attention (blocked), video-to-audio attention (blocked), and audio-to-video attention (blocked). Finally, the RoPE indices are shown in sub-figure (f).}
\label{fig:config}
\end{figure*}
Our goal is to effectively inject an audio signal and an identity/scene reference image into a video diffusion transformer to generate a talking character video. Our framework, ICAC, is based on the FullDiT \cite{ju2025fulldit}, as shown in Figure \ref{fig:architecture}. The core of our investigation lies in how the audio tokens $\boldsymbol{c}_\textrm{audio}\in\mathbb{R}^{B\times C\times F\times L}$, noisy video tokens $\boldsymbol{x}_t\in\mathbb{R}^{B\times C\times F\times N}$ and other asynchronous signals like image conditions $\boldsymbol{y}\in\mathbb{R}^{B\times C \times N}$ are processed within the transformer blocks, where $B$, $C$, $F$, $L$, $N$ denote the batch size, the channel number, the frame number, the audio spatial length, and the video spatial length, respectively. We explore three architectural variants, including 2D cross-attention, 2D self-attention, and unified 3D attention. Among these variants, we separate the 2D self-attention into with or without updating audio features, and study the attention masks for unified 3D attention.

\subsection{Architectural Exploration for Audio Injection}

We consider the following four configurations, each subsequent configuration possessing higher degrees of freedom. Intuitively, this progression corresponds to increasing model capacity as well. The overview of each configuration is shown in Figure \ref{fig:config} (a)-(d).

\textbf{(a) 2D cross-attention.} This approach follows the conventional method for adding conditioning to diffusion models~\cite{wei2025mocha,cui2025hallo3,ji2025sonic,jiang2024loopy,meng2025echomimicv2}. We add an extra 2D cross-attention layer into each transformer block, where the video tokens $\boldsymbol{x}_t$ act as the query and the audio tokens act as the key and value. The operation is done within the spatial dimension, \textit{i.e.}, between $L$ and $N$. As a result, this strategy insures strict frame alignment between video and audio. While simple and effective, this separates the processing of video and audio streams, potentially limiting the depth of their interaction.

\textbf{(b) 2D self-attention without updating audio features.} Compared to the 2D cross-attention baseline, this variant allows video tokens also attend to themselves, which increases the flexibility to some extent. At each block, we add an extra 2D self-attention layer, whose input is the concatenation between audio tokens corresponding to that frame and the spatial video tokens for that frame. At the attention output, we only maintain the updated video tokens, while drop the updated audio tokens and always use the initial audio tokens.

\textbf{(c) 2D self-attention.} Based on (b), after attention we also update the audio tokens. In this case, audio tokens will be updated according to their attention to both video and audio tokens. Similar to the FullDiT framework, this architecture treats audio as just another `in-context' condition.

\textbf{(d) Unified 3D self-attention.} This architecture represents the most integrated approach. The entire sequence of video tokens $\boldsymbol{x}_t\in \mathbb{R}^{B\times C\times (F\times N)}$, the sequence of other conditions $\boldsymbol{c}_\textrm{other}\in \mathbb{R}^{B\times C\times (F_\textrm{other} \times N)}$ (suppose other conditional tokens has the same spatial tokens with video but different frame numbers, such as image conditions) and the corresponding sequence of audio tokens $\boldsymbol{c}_\textrm{audio}\in \mathbb{R}^{B\times C\times (F \times L)}$ are concatenated into a single, long sequence. This combined sequence is then processed by the existing full 3D (spatio-temporal) self-attention mechanism. This allows every token (both video and audio) to attend to every other token across both space and time, offering the highest potential for learning complex audio-visual correlations, such as how the rhythm of speech influences body motion over time.

\subsection{Masked 3D Attention for Synchronous Signals}

During our initial experiments, we find that training with configuration (d) is extremely challenging. The model struggled to converge, and the generated videos exhibited poor lip-sync. We hypothesize this is because the full attention mechanism is too unconstrained for a signal that has a strict temporal relationship with the video. A phoneme at time $t$ should only influence the mouth shape at or very near time $t$. A full 3D attention allows a video token at one frame to attend to audio from a distant frame, which is physically implausible and provides confusing learning signals. To address this, we introduce Masked 3D Attention. The core idea is to apply an attention mask that restricts the receptive field of tokens to maintain temporal locality between the audio and video streams. Specifically, for a video token, it can attend to other video tokens at any temporal position (to model motion) but can only attend to audio tokens at the same frame. Similarly, we limit the audio tokens to only attend to audio and video tokens within each frame. The mask structure is visualized in Figure~\ref{fig:config} (e): the video-to-audio, audio-to-video, audio-to-audio sub-matrices are sparse and banded along the diagonal, enforcing that attention is localized in time.

\textbf{Efficient implementation with Flash Attention.} 
Flash Attention \cite{dao2022flashattention,dao2023flashattention2} offers a highly optimized attention implementation, but the off-the-shelf version does not support our custom block-banded attention mask. To maintain training efficiency, we provide an efficient implementation using Flash Attention. This allows us to train the large-scale 3D attention model efficiently, achieving significant speedups and memory savings compared to a na\"ve implementation with PyTorch's scaled dot product attention. Our approach implements the masked attention by decomposing the sparse attention matrix into a series of dense or block-diagonal sub-problems that can be efficiently computed by the standard Flash Attention kernel. The partial results from these computations are then merged using the online softmax trick, leveraging the LogSumExp (LSE) statistic that Flash Attention can optionally return. The scaled dot-product attention is defined as:
\begin{equation}
    \boldsymbol{O} = \text{softmax}\left(\frac{\boldsymbol{Q}\boldsymbol{K}^T}{\sqrt{d_k}}\right)\boldsymbol{V},
\end{equation}
where $\boldsymbol{Q}, \boldsymbol{K}, \boldsymbol{V}$ are the query, key, and value matrices, and $d_k$ is the dimension of the keys. Flash attention computes the output $\boldsymbol{O}$ without materializing the large intermediate $\boldsymbol{Q}\boldsymbol{K}^T$ matrix. Crucially for our method, it can also return the LogSumExp (LSE) of the unnormalized attention scores for each query row $i$:
\begin{equation}
    \text{LSE}_i = \log\sum_j \exp\left(s_{ij}\right), \quad \text{where} \quad s_{ij} = \frac{\boldsymbol{q}_i \cdot \boldsymbol{k}_j}{\sqrt{d_k}}
\end{equation}

Given two partial attention outputs, $(\boldsymbol{O}_1, \text{LSE}_1)$ and $(\boldsymbol{O}_2, \text{LSE}_2)$, computed over two disjoint sets of keys for the same queries, they can be merged into a final, correctly normalized output. This is achieved by first combining the LSEs:
\begin{equation}
    \begin{split}
        \text{LSE} &= \text{logaddexp}(\text{LSE}_1, \text{LSE}_2) \\
        &= \log(\exp(\text{LSE}_1) + \exp(\text{LSE}_2))
    \end{split}
\end{equation}
Then, the final output $\boldsymbol{O}$ is computed as a weighted sum of the partial outputs, where the weights are derived from the LSEs:
\begin{equation}
    \boldsymbol{O} = e^{(\text{LSE}_1 - \text{LSE})} \boldsymbol{O}_1 + e^{(\text{LSE}_2 - \text{LSE})} \boldsymbol{O}_2
\end{equation}

The entire process is summarized in Algorithm \ref{alg:masked_flash_attention}, where we apply Flash Attention multiple times to handle the customized mask.

\begin{algorithm}\small
\caption{Masked 3D Attention with Flash Attention}
\label{alg:masked_flash_attention}
\begin{algorithmic}
\STATE \textbf{Input:} Query $\boldsymbol{Q}$, Key $\boldsymbol{K}$, Value $\boldsymbol{V}$
\STATE \textbf{Output:} Attention output $\boldsymbol{O}$

\STATE // Partition matrices based on modality (video, others, audio) and attention pattern
\STATE $\boldsymbol{Q} \to (\boldsymbol{Q}_\text{video}, \boldsymbol{Q}_\text{others}, \boldsymbol{Q}_\text{audio})$
\STATE $\boldsymbol{K} \to (\boldsymbol{K}_\text{video}, \boldsymbol{K}_\text{others},\boldsymbol{K}_\text{audio})$
\STATE $\boldsymbol{V} \to (\boldsymbol{V}_\text{video}, \boldsymbol{V}_\text{others},\boldsymbol{V}_\text{audio})$
\STATE $\boldsymbol{K}_\text{v-o} \leftarrow \text{concat} (\boldsymbol{K}_\text{video}, \boldsymbol{K}_\text{others})$
\STATE $\boldsymbol{V}_\text{v-o} \leftarrow \text{concat}(\boldsymbol{V}_\text{video}, \boldsymbol{V}_\text{others})$

\STATE // Step 1: Process video queries
\STATE $(\boldsymbol{O}_{\text{v} \to \text{vo}}, \text{LSE}_{\text{v} \to \text{vo}}) \leftarrow \text{FlashAttention}(\boldsymbol{Q}_\text{video}, \boldsymbol{K}_\text{v-o}, \boldsymbol{V}_\text{v-o})$
\STATE $(\boldsymbol{O}_{\text{v} \to \text{a}}, \text{LSE}_{\text{v} \to \text{a}}) \leftarrow \text{FlashAttention}(\boldsymbol{Q}_\text{video}, \boldsymbol{K}_\text{audio}, \boldsymbol{V}_\text{audio})$ \COMMENT{Uses `cu\_seqlens'}
\STATE $\text{LSE}_\text{video} \leftarrow \text{logaddexp}(\text{LSE}_{\text{v} \to \text{vo}}, \text{LSE}_{\text{v} \to \text{a}})$
\STATE $\boldsymbol{w}_1 \leftarrow \exp(\text{LSE}_{\text{v} \to \text{vo}} - \text{LSE}_\text{video})$; \quad $\boldsymbol{w}_2 \leftarrow \exp(\text{LSE}_{\text{v} \to \text{a}} - \text{LSE}_\text{video})$
\STATE $\boldsymbol{O}_\text{video} \leftarrow \boldsymbol{w}_1 \cdot \boldsymbol{O}_{\text{v} \to \text{vo}} + \boldsymbol{w}_2 \cdot \boldsymbol{O}_{\text{v} \to \text{a}}$

\STATE // Step 2: Process others queries
\STATE $(\boldsymbol{O}_\text{others}, \_) \leftarrow \text{FlashAttention}(\boldsymbol{Q}_\text{others}, \boldsymbol{K}_\text{v-o}, \boldsymbol{V}_\text{v-o})$

\STATE // Step 3: Process audio queries
\STATE $(\boldsymbol{O}_{\text{a} \to \text{v}}, \text{LSE}_{\text{a} \to \text{v}}) \leftarrow \text{FlashAttention}(\boldsymbol{Q}_\text{audio}, \boldsymbol{K}_\text{video}, \boldsymbol{V}_\text{video})$ \COMMENT{Uses `cu\_seqlens'}
\STATE $(\boldsymbol{O}_{\text{a} \to \text{a}}, \text{LSE}_{\text{a} \to \text{a}}) \leftarrow \text{FlashAttention}(\boldsymbol{Q}_\text{audio}, \boldsymbol{K}_\text{audio}, \boldsymbol{V}_\text{audio})$ \COMMENT{Uses `cu\_seqlens'}
\STATE $\text{LSE}_\text{audio} \leftarrow \text{logaddexp}(\text{LSE}_{\text{a} \to \text{v}}, \text{LSE}_{\text{a} \to \text{a}})$
\STATE $\boldsymbol{w}_1 \leftarrow \exp(\text{LSE}_{\text{a} \to \text{v}} - \text{LSE}_\text{audio})$; \quad $\boldsymbol{w}_2 \leftarrow \exp(\text{LSE}_{\text{a} \to \text{a}} - \text{LSE}_\text{audio})$
\STATE $\boldsymbol{O}_\text{audio} \leftarrow \boldsymbol{w}_1 \cdot \boldsymbol{O}_{\text{a} \to \text{v}} + \boldsymbol{w}_2 \cdot \boldsymbol{O}_{\text{a} \to \text{a}}$

\STATE // Step 4: Concatenate results
\STATE $\boldsymbol{O} \leftarrow \text{concat}(\boldsymbol{O}_\text{video}, \boldsymbol{O}_\text{others}, \boldsymbol{O}_\text{audio})$
\STATE \textbf{return} $\boldsymbol{O}$
\end{algorithmic}
\end{algorithm}

\subsection{Applying Rotary Position Embedding}
Rotary Position Embedding (RoPE) \cite{su2024roformer} is crucial in transformers, especially for synchronized modalities. It injects a sense of order into the otherwise unordered set of tokens that transformers process. While RoPE is initially designed for 1D sequences like text, its principles have been extended to handle 2D data, where the embedding vector is split, with one part encoding the $x$-coordinate and the other encoding the $y$-coordinate.
However, in our model we need to process the interleaved 2D audio, 2D image and 3D video data in a single concatenated sequence. To address this, a more elegant solution is to unify both text and image positions into a single 2D coordinate system \cite{kexuefm10040}. This approach not only preserves the vital 2D structure of images but is also designed to be backward-compatible with RoPE-1D. The RoPE indices for video and audio are shown in Figure \ref{fig:config} (f), where the shared indices for frame dimension are omitted, and only the indices for the spatial dimensions of video and audio are displayed:
\begin{itemize}
    \item Audio as a diagonal in 2D space: Each audio token at a 1D position $n$ is mapped to a 2D coordinate $(n, n)$.
    \item Video as inlaid 2D grids: The video represented as a grid of patches is na\"ively positioned in the 2D space.
\end{itemize}

\section{Experiments}
\label{sec:experiments}
\begin{figure*}[t!]
\begin{center}
\includegraphics[width=\linewidth]{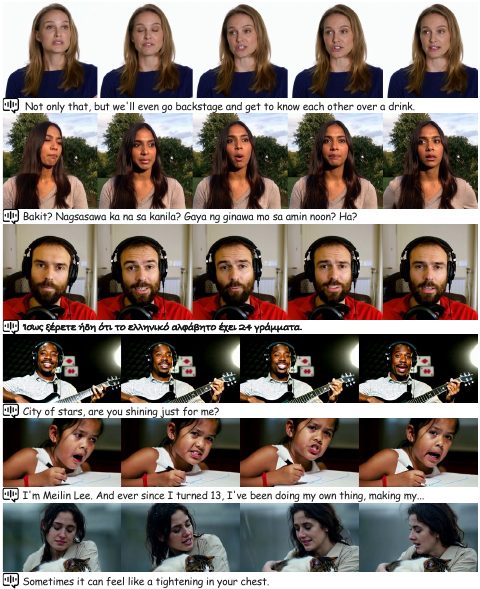}
\end{center}
\caption{Generated results from Celeb-V (the first three rows) and MochaBench (the rest).}
\label{fig:visual}
\end{figure*}
\subsection{Implementation Details}
This section details the architecture of the model used, the datasets employed for training and evaluation, the metrics for quantitative assessment, and the training configurations.

\noindent\textbf{Model and initialization.} Our model, ICAC, is built upon the FullDiT-1B architecture \cite{ju2025fulldit}. We initialize our training from an internal checkpoint pre-trained for multi-identity image-to-video synthesis. This provides a strong foundation for visual and identity representation. We extract the audio feature using Wav2Vec2 \cite{baevski2020wav2vec} as an additional control signal and finetune the entire model.

\noindent\textbf{Training dataset.} For training our model, we utilize the comprehensive dataset curated and introduced in Hallo3 \cite{cui2025hallo3}. This dataset is specifically designed for high-quality talking portrait animation and is composed of approximately $134$ hours of video data sourced from three distinct domains to ensure diversity and generalization. The composition includes $6$ hours from the HDTF dataset, $72$ hours from public YouTube videos, and $56$ hours from a large-scale movie dataset. The data underwent a meticulous curation pipeline involving single-speaker extraction, motion filtering, and post-processing to ensure high quality and relevance for audio-driven facial animation.

\noindent\textbf{Evaluation benchmarks.} To assess the performance of our model, we conduct evaluations on two distinct benchmarks known for their relevance to talking character generation and their challenging, in-the-wild characteristics.
\begin{itemize}
    \item MoCha-Bench \cite{wei2025mocha}. It consists of $150$ diverse examples, each with a textual prompt and a corresponding audio clip. The benchmark includes both close-up shots that emphasize facial expressions and medium shots that highlight hand gestures and body movements, providing a comprehensive test of a model's ability to generate realistic and contextually appropriate animations.
    \item Celeb-V \cite{zhu2022celebv}. As a widely recognized benchmark in the field, Celeb-V is a large-scale, high-quality video dataset of celebrities in various real-world settings. It is used to evaluate the performance in generating realistic lip movements and consistent identities for subjects in the wild. We select $100$ samples where the characters and speech are clearly distinguishable to form the test set.
\end{itemize}

\noindent\textbf{Evaluation metrics.} Our quantitative analysis focuses on lip-synchronization accuracy, a critical aspect of talking portrait generation. To this end, we employ two standard, pre-trained metrics derived from the SyncNet model \cite{chung2016out}, including Sync-C (confidence Score, higher is better) \cite{chung2016out} and Sync-D (distance, lower is better) \cite{chung2016out}.
\begin{table*}[t!]
\caption{Comparison of audio injection architectures on MochaBench \cite{wei2025mocha}.}
\label{tab:ablation}
\centering
\addtolength{\tabcolsep}{3pt}
\begin{tabular}{lcccc}
\toprule
Architecture & Configuration & Convergence Stability & Sync-C $\uparrow$ & Sync-D $\downarrow$ \\
\midrule
2D Cross-Attention & Config (a) & Fast & 3.199 & 10.236 \\
Partial 2D Self-Attention & Config (b) & Moderate & 3.301 & 9.993 \\
Full 2D Self-Attention & Config (c) & Moderate & 3.702 & 9.852 \\
3D Self-Attention & Config (d) (w/o mask (e)) & Failed to Converge & 1.562 & 12.291 \\
Masked 3D Self-Attention \textbf{(Ours)} & Config (d) (w/ mask (e)) & Stable & \textbf{4.826} & \textbf{9.121} \\
\bottomrule
\end{tabular}
\end{table*}

\noindent\textbf{Training configurations.} We conduct all training runs with a total effective batch size of $128$. A constant learning rate of $1e-5$ is used for all experiments. The training strategies for our architectural variants differ based on their complexity. For the configurations that do not employ unified 3D attention, a direct, single-stage training procedure is sufficient. These models are trained with both audio and image conditions provided simultaneously from the beginning. We observe that they converge reliably, and their performance plateau after approximately $12500$ training iterations. In contrast, training the unified 3D self-attention model presents a unique challenge. When audio and image conditions are introduced simultaneously, we find that the powerful, static features from the identity/scene image tend to dominate the training process. This makes it difficult for the model to effectively learn from the more subtle, dynamic audio signal, leading to poor injection of the audio condition and inadequate lip-sync. To mitigate this issue, we use a two-stage training curriculum:
\begin{itemize}
\item \textbf{Stage 1: Audio-Only Pre-training.} We first train the model for $12500$ iterations using only the audio condition. This initial phase force the network to focus exclusively on learning the mapping from audio features to visual speech dynamics without the overpowering influence of the image.
\item \textbf{Stage 2: Joint Finetuning.} Following the first stage, we introduce the image condition and continue training for an additional $12500$ iterations. In this stage, the model, already adept at handling audio, learns to integrate the identity and scene information while preserving the audio-visual synchronization.
\end{itemize}
After this two-stage process, the performance of the 3D attention model also reaches a plateau, indicating that it has fully converged.

\begin{table}[t!]\small
\caption{Performance comparison with other methods on Celeb-V \citep{zhu2022celebv}.}
\label{tab:celebv}
\centering
\addtolength{\tabcolsep}{-1.2pt}
\begin{tabular}{llccccc}
\toprule
\textbf{Method} & \textbf{Arch.} & \textbf{\#Params} & \textbf{Sync-C $\uparrow$} & \textbf{Sync-D $\downarrow$} \\
\midrule
EchoMimic \cite{chen2025echomimic} & U-Net & 1.7B & 4.072 & 10.092 \\
Hallo3 \cite{cui2025hallo3} & DiT & 5B & 5.206 & 9.352 \\
Sonic \cite{ji2025sonic} & U-Net & 1.5B & 6.939 & 7.461 \\
\textbf{ICAC (Ours)} & FullDiT & 1B & 5.002 & 9.123 \\
\bottomrule
\end{tabular}
\end{table}

\begin{table}[t!]\small
\caption{Performance comparison with other methods on MochaBench \citep{wei2025mocha}.}
\label{tab:mochabench}
\centering
\addtolength{\tabcolsep}{-1.5pt}
\begin{tabular}{llccccc}
\toprule
\textbf{Method} & \textbf{Arch.} & \textbf{\#Params} & \textbf{Sync-C $\uparrow$} & \textbf{Sync-D $\downarrow$} \\
\midrule
SadTalker \cite{zhang2023sadtalker} & - & - & 4.727 & 9.239 \\
AnyPortrait \cite{wei2024aniportrait} & U-Net & 2B & 1.740 & 11.383 \\
Hallo3 \cite{cui2025hallo3} & DiT & 5B & 4.866 & 8.963 \\
MoCha \cite{wei2025mocha} & DiT & 30B & 6.037 & 8.103 \\
Sonic \cite{ji2025sonic} & U-Net & 1.5B & 6.842 & 7.440 \\
\textbf{ICAC (Ours)} & FullDiT & 1B & 4.826 & 9.121 \\
\bottomrule
\end{tabular}
\end{table}

\subsection{Ablation Study on Audio Injection}

To validate our architectural choices, we conduct comparison experiments of four aforementioned configurations for injecting the audio signal into the unified transformer framework. We evaluate each method based on its training stability (convergence) and final performance on key metrics. As shown in Table \ref{tab:ablation}, our findings reveal a clear trade-off. The cross-attention model (a) converges the fastest but yields the lowest performance. The 2D self-attention model (b) and (c) converges slightly slower but achieves better results. The unconstrained unified 3D self-attention (d), while theoretically the most powerful, failed to converge on the lip-sync task, resulting in unstable and incoherent motion. However, our proposed Masked 3D Attention not only converges stably but also significantly outperforms all other methods, confirming that constraining the spatio-temporal attention is crucial for unlocking the full potential of a unified architecture for synchronous signals.

\subsection{Comparison with State-of-the-Art Methods}

We compare our final ICAC model against several leading audio-driven video generation methods. As shown in Table \ref{tab:celebv}, on the Celeb-V benchmark, ICAC achieves highly competitive performance, with lip-sync scores that rival top methods. On the more diverse MochaBench (Table \ref{tab:mochabench}), our model shows strong generalization capabilities. The significance of these results becomes clear when comparing directly with the much larger Hallo3 model: trained on the exact same Hallo3 audio dataset, our lean 1B parameter model consistently delivers performance comparable to their 5B model. This is especially true on MochaBench, where our model overcomes the domain gap without any in-domain fine-tuning. This achievement highlights the power and efficiency of our Masked 3D Attention mechanism within a unified, in-context framework. The visualizations in Figure \ref{fig:visual} further confirm our model's high-quality output and accurate lip synchronization.

\section{Conclusion}
In this work, we address the fundamental challenge of integrating a time-synchronous signal, such as audio, into a unified in-context video generation transformer. Our systematic investigation reveals that while a fully unified 3D self-attention approach is powerful in nature, it is plagued by significant convergence issues. To resolve this, we propose the Masked 3D Attention mechanism, an effective solution that stabilizes training by enforcing local temporal alignment between audio and video features. Embodying this strategy, our final model is able to generate high-quality, well-synchronized talking character videos, offering a promising blueprint for future multi-modal transformer architectures.
{
    \small
    \bibliographystyle{ieeenat_fullname}
    \bibliography{main}
}


\end{document}